\documentclass{article}

\PassOptionsToPackage{numbers,compress}{natbib}

\usepackage[dblblindworkshop,final]{neurips_2025}

\workshoptitle{Learning from Time Series for Health (TS4H)}

\usepackage[utf8]{inputenc} 
\usepackage[T1]{fontenc}    
\usepackage{hyperref}       
\usepackage{url}            
\usepackage{booktabs}       
\usepackage{amsfonts}       
\usepackage{amsmath}
\usepackage{amssymb}
\usepackage{nicefrac}       
\usepackage{microtype}      
\usepackage{xcolor}         
\usepackage[capitalize]{cleveref}
\usepackage{circledsteps}
\usepackage{comment}
\usepackage{graphicx}
\usepackage{xcolor}
\usepackage{subcaption}
\usepackage{multirow}
\usepackage{wrapfig}
\usepackage{bm} 
\usepackage{placeins}

\definecolor{pltTeal}{RGB}{85,175,169}
\definecolor{pltMagenta}{RGB}{233,52,247}
\definecolor{pltOrange}{RGB}{242,169,59}

\title{Improving Forecasts of Suicide Attempts \\ for Patients with Little Data}

\author{%
  Genesis Hang \\
  Wellesley College\\
  \texttt{gh104@wellesley.edu} \\
  \And
  Annie Chen \\
  Wellesley College\\
  \texttt{ac134@wellesley.edu} \\
  \And  
  Hope Neveux \\
  Harvard University \\
  \texttt{hopeneveux@fas.harvard.edu} \\
  \AND
  Matthew K.~Nock \\
  Harvard University \\
  \texttt{nock@wjh.harvard.edu} \\
  \And
  Yaniv Yacoby \\
  Wellesley College\\
  \texttt{yy109@wellesley.edu} \\
}

\begin{document}

\maketitle

\begin{abstract}
Ecological Momentary Assessment (EMA) studies provide real-time data on suicidal thoughts and behaviors, but forecasting suicide attempts remains challenging: attempts are rare, and the pathways patients take to them are heterogeneous. Here, we investigate a cohort of patients from an EMA study with recorded suicide-related events. We show that a single model fit to all patients forecasts poorly, while idiographic (per-patient) models show improvement but overfit for those with little data. Based on this result, one may hypothesize that patients should be partitioned into subgroups---this way, similar patients' data can be pooled together to improve forecasts. However, we show that grouping patients at random already improves forecasts, with performance increasing monotonically with the number of groups. Moreover, we show that grouping patients by demographics yields worse forecasts than random groupings. From these results, we hypothesize that patient similarity is continuous, rather than discrete, and must be inferred from the data. This motivated us to use Latent Variable Multiple Output Gaussian Processes (LVMOGPs), adapted to our data. Preliminary results show that, even without careful kernel design, LVMOGPs already match the strongest baseline models on most metrics, and their latent spaces yield a similarity between patients that we can inspect directly. Because the cohort is conditioned on the outcome and the splits are not temporal, we read these results as evidence that idiographic structure exists and can be recovered, not as deployable forecasting performance---an area for future work.
\end{abstract}

\section{Introduction and Related Work} 

Ecological Momentary Assessment (EMA) studies leverage smartphones to capture insights into suicidal thoughts and behaviors (STBs) as they unfold in daily life~\cite{shiffman2008ecological,nock2010revealing,nock2026understanding}.
In these intensive longitudinal studies, patients are surveyed multiple times a day about their suicidal urges, intent, and affects.
The sampling rate in this dataset presents opportunities for machine learning (ML) to forecast imminent suicide risk in time for intervention; however, to date, no current approach can do this reliably~\cite{kleiman2023use}.

Prior work primarily focuses on forecasting suicidal ideation from EMA data (e.g.~\cite{kleiman2017examination,czyz2023ecological,lei2023ecological,wang2023idiographic}).
While forecasting ideation is itself challenging, suicide attempts are even harder to predict due to their low base rate~\cite{fox2020interventions}; even in the largest available datasets (e.g.~600 patients), attempts are rarely captured (e.g.~\cite{czyz2018ecological}). 
This severely limits the data available for model training and evaluation.
Exacerbating this challenge, recent work shows that patients' paths to suicidal ideation are heterogeneous, suggesting that, at the very least, there are many subtypes of at-risk patients, and thus constitutes evidence against forecasting with a single predictive trend for all patients~\cite{kleiman2018digital,kaurin2022integrating,coppersmith2024heterogeneity}.
But training independent trends for different patient subgroups further reduces the number of training data points available per trend, and risks drastically reducing the forecasting performance for patients with little data.

In this work, we show that the same patient heterogeneity found in the prediction of suicidal ideation is also found in the prediction of suicide attempts. 
We then present a model class that is able to leverage information from all patients and simultaneously improve forecasts for patients with little data by learning the structure of patient heterogeneity.
Our contributions are:

\textbf{(A) We show that a single predictive trend trained on data from all patients to forecast suicide attempts performs worse than idiographic models.}
We construct a cohort of patients who \emph{do} have recorded suicide-related events (SREs), keeping only those with at least three positively- and three negatively-labeled surveys, so that every data split contains both classes (\cref{apx:setup}).
Then, we show that each patient exhibits different predictive trends that, when combined, conflict with one another, resulting in poor forecasting performance.
This underscores the importance of explicitly modeling patient heterogeneity~\cite{coppersmith2024heterogeneity,kaurin2022integrating}.  

From these results, we may be tempted to use a different predictive trend per patient, but we show that idiographic models are prone to overfitting on patients with little data.
Moreover, they require us to collect enough data per patient to make informed forecasts, limiting their clinical utility. 

\textbf{(B) We show that an improvement from grouping patients is not, by itself, evidence that the groups are meaningful.}
A natural way to address patient heterogeneity is to partition patients into subgroups and fit a predictive trend per subgroup. This way, similar patients' data can be pooled together to improve forecasts. We find that assigning patients to $G$ groups \emph{completely at random} already improves forecasts across nearly all metrics, with the improvement increasing with $G$. The improvement from random subgrouping has broader implications: a forecasting gain from modeling with subgroups does not automatically constitute evidence that the subgroups have captured meaningful structures in the data. In fact, we find that grouping by patient demographics available to us performs \emph{worse} than grouping at random, suggesting these demographic labels alone may not capture the axes of patient heterogeneity. Based on these results, we propose that heterogeneity should be represented as a latent variable in the model, to be inferred from the data. 

\textbf{(C) We translate dataset features into model class requirements and propose a model class with the desired property.}
Taken together, (A) and (B) suggest a model in which patient heterogeneity is modeled as a \emph{continuous latent variable} so that patients with little data can borrow statistical strength from similar patients. To satisfy this criterion, we propose a type of Latent Variable Multiple Output Gaussian Process (LVMOGP) \citep{dai2017efficient}. Because SREs are binary while LVMOGP is a regression model, we replace its Gaussian likelihood with a Bernoulli one and fit the result by sparse variational inference. We show that LVMOGP matches the strongest baselines on most metrics, even without additional effort in kernel design or hyperparameter search. 

\textbf{(D) We propose a new way of deriving patient similarity from the learned latent space of our model. We show that the patient similarity we learn accounts for heterogeneity in the dataset, and that the learned similarity is not accounted for by patient demographics provided to us.}
The fitted latent kernel yields a similarity between every pair of \emph{patients}, which we treat as a weighted graph: we visualize it, color it by demographic group, and quantify how far similarity aligns with a grouping. This turns a latent space from a mere means for prediction into a scientific object of interest. 

\begin{figure*}
    \centering
    
    \begin{subfigure}{0.34\textwidth}
    \includegraphics[width=\linewidth]{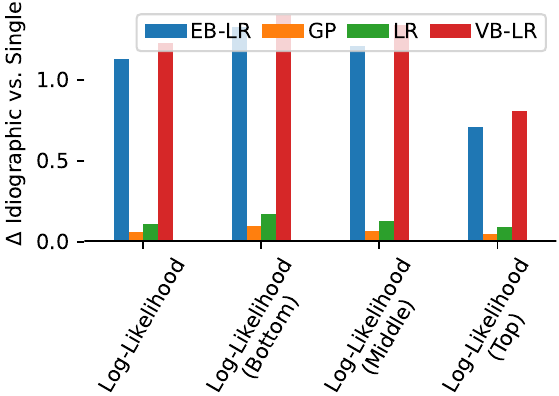}
    \end{subfigure}
    ~
    \begin{subfigure}{0.64\textwidth}
    \includegraphics[width=\linewidth]{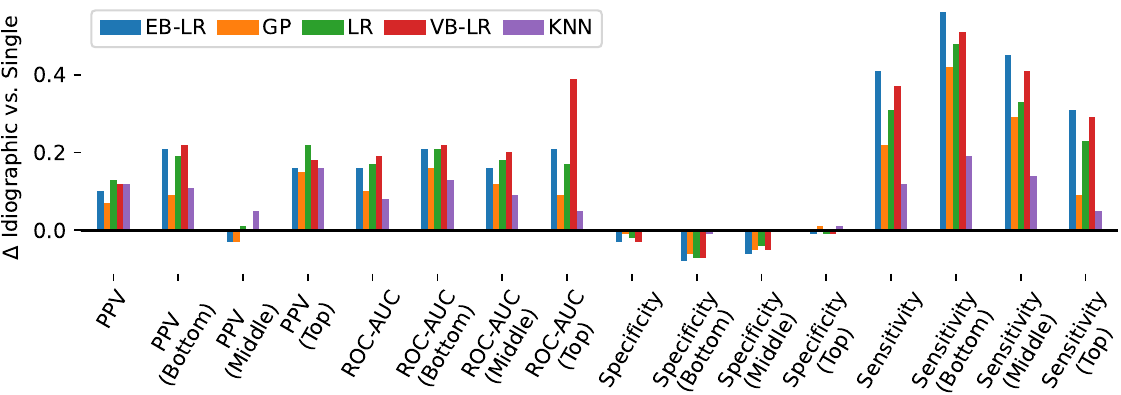}
    \end{subfigure}

    \vspace{1mm}
    \hrule
    \vspace{1mm}

    \begin{subfigure}{0.46\textwidth}
    \includegraphics[width=\linewidth]{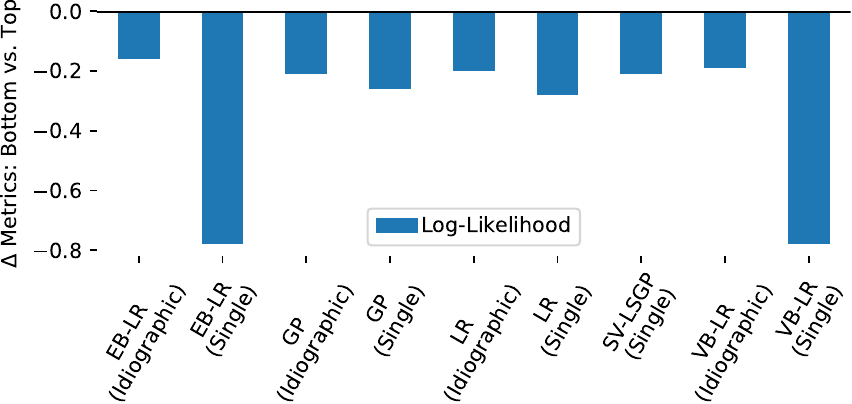}
    \end{subfigure}
    ~
    \begin{subfigure}{0.52\textwidth}
    \includegraphics[width=\linewidth]{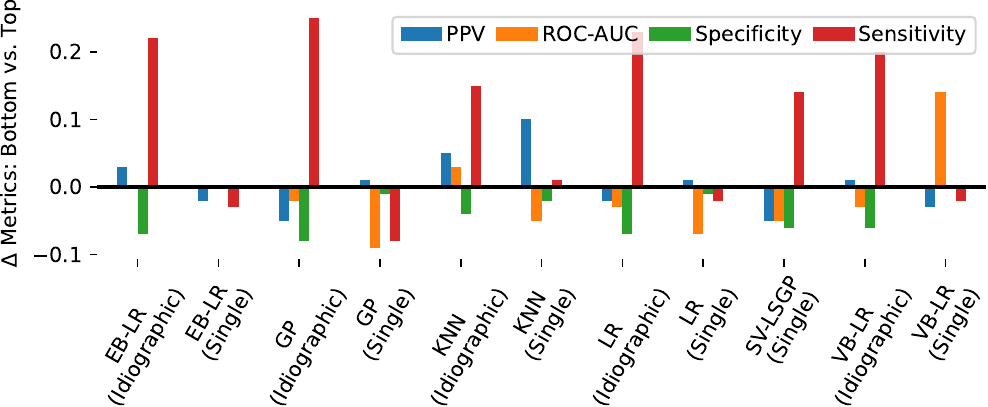}
    \end{subfigure}
    
    \caption{\textbf{Top Row: Idiographic models outperform their Single counterparts}---except for specificity, which decreased slightly, the magnitude of difference in metrics between the idiographic and single models is almost always positive. \textbf{Bottom Row: The 30\% of patients with the fewest data points consistently receive worse forecasts across most metrics and most models than the 30\% of patients with the most data}---except for sensitivity, the magnitude of difference in metrics computed is usually negative, indicating lower performance.}
    \label{fig:comparisons}
\end{figure*}

\section{Forecasts for At-Risk Patients Require Capturing Heterogeneity} \label{sec:geometry}

In a series of experiments, we show that popular modeling approaches fail to both capture patient heterogeneity and preserve predictive performance, motivating our proposed modeling approach.

\textbf{Notation.}
Let $N$ denote the number of patients in the data.
Let $x_i = \begin{bmatrix} r_{i,1} & \dots & r_{i,D_x} \end{bmatrix}^\intercal$ represent the $i$th observation in the data, belonging to patient $n_i$ at time $t_i$, consisting of their responses $r_{i,d} \in \{ 0, \dots, 10 \}$ to 0--10 Likert-scale EMA questions.
Here, we will use questions about patients' affects and their suicidal urge, intent, and ability to resist urges (details in \cref{apx:data}).
Using patient responses to these questions, our task is to predict $y_i \in \{ 0, 1 \}$---whether patient $n_i$ engaged in any suicide-related event (SRE) sometime in the week following $t_i$.
We define an SRE as either a self-injurious behavior with some (non-zero) intention of dying, or a presentation to a hospital with suicidal thoughts to prevent a suicide attempt. 
Let $\mathcal{D} = X, Y$ represent the entire training data.
Let $\mathcal{D}_n = X_n, Y_n$ represent patient $n$'s training data, where $X_n = \{ x_i | n_i = n \}$ and $Y_n = \{ y_i | n_i = n \}$.
Note that every patient has a different amount of data.

\textbf{Goal.}
Given $\mathcal{D}$, our goal is to predict whether, given a \emph{new} EMA response, $x_n^*$, patient $n$ will engage in an SRE sometime in the next week, $y_n^*$.

\textbf{Single vs.~Idiographic Models.}
To better understand the geometry of patient classification boundaries, we compare a model trained on \emph{all} patient data ($y_n^* | x_n^*, \mathcal{D}$) with a model consisting of a collection of models---\emph{one per patient} ($y_n^* | x_n^*, \mathcal{D}_n$).
We refer to the former and latter as a single and an idiographic model, respectively.
If idiographic models consistently outperform the single models, this suggests that patients have differing forecasting trends.
Even before comparing their performance, we note that idiographic models have one major shortcoming: they cannot be used to make predictions for a new patient $n^*$; we address this limitation in \cref{sec:method}. 

\textbf{Baselines and Metrics.}
We consider several popular and natural baselines, each used both as a single and as an idiographic model: Gaussian Process Classification with an isotropic RBF kernel (RBF-GP), k-Nearest Neighbor Classifier (KNN), Logistic Regression (LR), and Bayesian LR with empirical Bayes Type II and variational (EB-LR and VB-LR) approximations.
For evaluation, we use: Avg.~Log-Likelihood, Positive Predictive Value (PPV), Area Under the Receiver Operating Characteristic Curve (ROC-AUC), Specificity, and Sensitivity. 

\begin{figure*}
    \centering
    
    \begin{subfigure}{1.0\textwidth}
    \includegraphics[width=\linewidth]{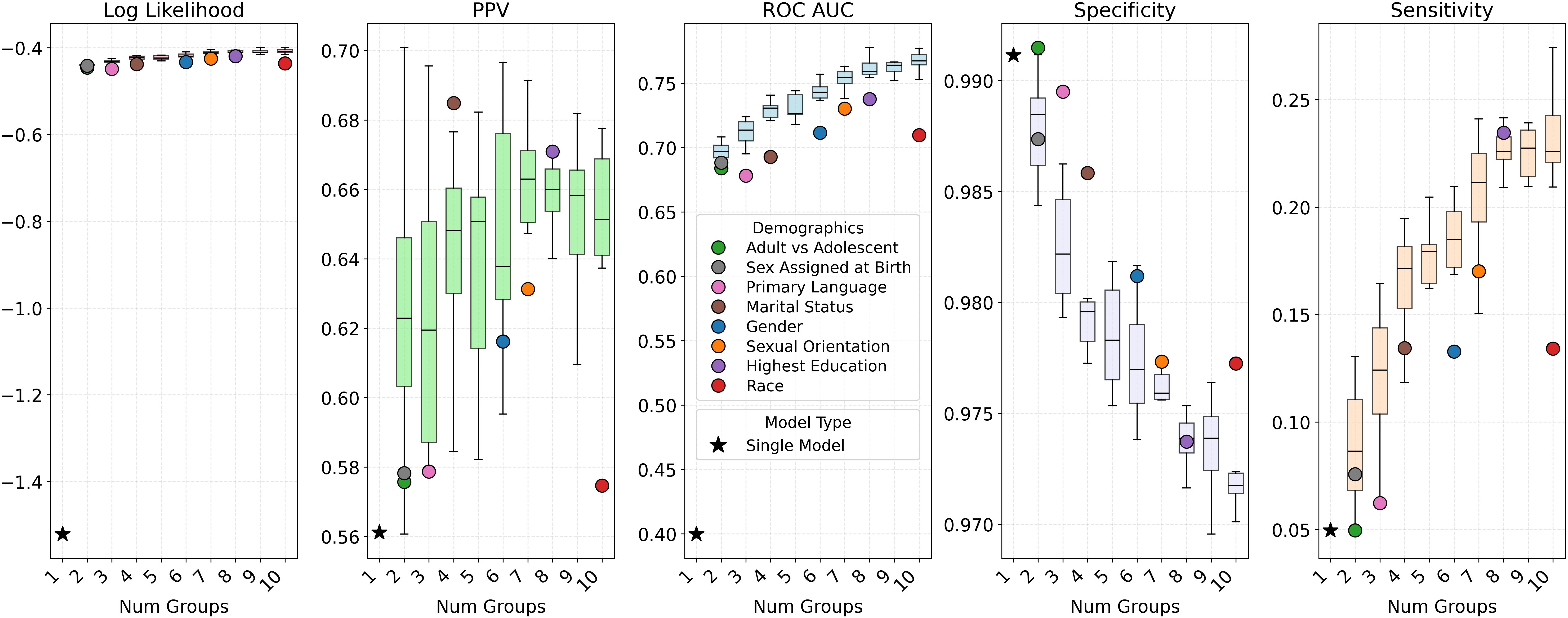}
    \end{subfigure}
    
    \caption{\textbf{Patient heterogeneity is so high that even random groupings of patients significantly boost performance.} We split the dataset into $G$ groups, randomly assigning patients, fitting a separate model to each group, and measuring the performance (y-axis) as $G$ increases (x-axis). We repeated this experiment 10 times per $G$, plotting the distribution of metrics. Finally, we also compare models fit on randomly- vs.~demographics-grouped patients (using the same number of groups, $G$).}
    \label{fig:random-groups-experiment}
\end{figure*}

\textbf{Finding 1: Patients exhibit heterogeneity, i.e.~conflicting classification boundaries.}
\cref{fig:comparisons} (top row) shows that, across all baselines and metrics (except specificity), idiographic models outperform their single counterparts. 
As such, we may be tempted to just use separate models; however, as \cref{fig:comparisons} (bottom row) shows, both single and idiographic models make worse forecasts for the patients with less data.
Single model predictions are most influenced by patients with more data, generalizing poorly to patients with less data; idiographic models overfit to patients with little data.

\textbf{Finding 2: Discrete groups may fail to account for patient heterogeneity.}
Given the heterogeneity of forecasting trends, we may want to find \emph{discrete groupings} of patients with similar trends.
However, as we show here, our patient trends are so heterogeneous that \emph{even randomly grouping patients significantly boosts performance}, increasing the risk of mis-grouping patients and misleading scientific conclusions. 
To show this, we conduct a simple experiment: we split the dataset into $G$ groups by \emph{randomly} assigning each patient's data to a group, fitting a separate model to each group, and measuring the performance of the models as $G$ increases.
As \cref{fig:random-groups-experiment} shows, with increasing $G$, all metrics improve (except for specificity, which decreases a little to increase sensitivity).
Even more interestingly, models trained on patients grouped by demographics tend to perform worse than those trained on randomly grouped patients on nearly all metrics.
This leads us to question whether our demographic data explain patient heterogeneity.
We next turn to a model that captures a \emph{continuous} notion of patient similarity.

\section{Modeling Patient Similarity as a Continuous Latent Variable} \label{sec:method}

There are many pathways to suicide; even among mental health disorders, conditions such as major depression, generalized anxiety, post-traumatic stress, and borderline personality disorder each present distinct mechanisms leading to elevated suicide risk~\cite{hawton2000international,blasco2016patterns}. 
Moreover, within a single diagnosis, each patient's unique life circumstances---e.g.~shaped by social determinants and individual differences---may further contribute to patient heterogeneity~\cite{kaurin2022integrating}. 
To capture this probabilistically, we seek a model that interpolates between the single and idiographic models of \cref{sec:geometry}, accounting for each patient's individual forecasting trajectory while still enabling those with limited data to leverage data from others.
We address this by embedding patients in a latent space, where proximity reflects similarity in risk trajectories: patients with nearby codes share statistical strength, approaching the single model, while patients with dissimilar codes forecast independently, approaching the idiographic model. Forecasts for patients with little data can thus intelligently ``borrow trends'' from their neighbors. This is the construction of the LVMOGP, with the patient in the role of the condition.

\textbf{The Model Definition.}
The findings from \cref{sec:geometry} support our modeling approach---an LVMOGP model with two differences, a Bernoulli (instead of Gaussian) likelihood, and unconstrained (as opposed to Kronecker) inducing points:
\begin{align}
    &\Circled{A}\text{  } z_n \sim p(z) = \mathcal{N}(0, \mathbb{I}_{D_z}), & &\Circled{B}\text{  } \hat{x}_i = [ r_{i,1} \text{  } \dots \text{  } r_{i, D_x} \text{  } z_{n_i,1} \text{  } \dots \text{  } z_{n_i, D_z} ]^\intercal, \\ 
    &\Circled{C}\text{  } F | \widehat{X}; \theta \sim \mathcal{N}( 0, K_\theta(\widehat{X}, \widehat{X}) ), & &\Circled{D}\text{  } y_i | f_i \sim \mathrm{Bernoulli}( \mathrm{sigmoid}(f_i) ), 
\end{align}
where $\hat{x}_i$ represents the concatenation of the inputs $x_i$ with the latent variable $z_{n_i}$ corresponding to patient $n_i$, $\widehat{X}$ is a matrix consisting of all $\hat{x}_i$'s as rows, $K_\theta(\cdot, \cdot)$ is the kernel matrix computed on rows of its arguments with hyperparameters $\theta$ (a single point counts as one row, so $K_\theta(\hat{x}_i, A)$ is the row vector of its kernel values against the rows of $A$), $F$ is a concatenation of all function values $f_i$ corresponding to each $\hat{x}_i$, and $\mathbb{I}_{D_z}$ is an identity matrix of dimension $D_z$.

\textbf{Related Models.}
LVMOGPs bear similarity to several other models.
GPs with Latent Covariates~\cite{wang2012gaussian} and Covariate GP Latent Variable Models~\cite{martens2019decomposing} attach a latent variable to every observation rather than to every patient, so nothing is shared within a patient. Multi-Group GPs~\cite{li2025bayesian} attach one to each group, but the group is observed and discrete rather than inferred and continuous. Meta-Learning GPs~\cite{saemundsson2018meta} also bear strong similarity to LVMOGPs, but assume a control signal that EMA data does not provide. To our knowledge, none of this family has been applied to forecasting SREs from EMA data, nor used to infer patient similarity off the learned latent space.

\textbf{A Sparse Variational Formulation to Enable Tractable Inference.} 
Analytical inference is impossible due to the non-Gaussianity of the likelihood and the large number of observations ($14794$ from $N = 75$ patients), so we apply the sparse variational formulation of GPs~\cite{titsias2009variational} to our model, replacing $\Circled{C}$ above with:
\begin{align}
    &\Circled{C.1}\text{  } U; W, \theta \sim \mathcal{N}( 0, K_\theta(W, W) ), \\
    &\Circled{C.2}\text{  } F | U, X, Z; W, \theta \sim \mathcal{N}( \Psi \cdot U, K_\theta(\widehat{X}, \widehat{X}) - \Psi \cdot K_\theta(\widehat{X}, W)^\intercal ), 
\end{align}
where $\Psi = K_\theta(\widehat{X}, W) \cdot K_\theta(W, W)^{-1}$.
In this formulation, $W \in \mathbb{R}^{M \times (D_x + D_z)}$ is a matrix of $M$ inducing point locations used to ``summarize'' the training data, enabling more efficient inference.

\textbf{Stochastic Variational Inference (SVI) for Fast Inference.}
We learn the $W$, $\theta$, and $\phi$ that minimize the divergence between an approximate and the true posterior~\cite{titsias2009variational}:
\begin{align}
    W^*, \theta^*, \phi^* &= \mathrm{argmin}_{W, \theta, \phi} D_\text{KL} \left[ q(F, U, Z; W, \theta, \phi) || p(F, U, Z | X, Y; W, \theta) \right],
\end{align}
using the variational family, $q(F, U, Z; W, \theta, \phi) = 
p(F | U, X, Z; W, \theta) \cdot q(U; \phi) \cdot \prod_{n=1}^N q(z_n; \phi)$, where $\phi$ are the parameters of a full-covariance Gaussian $q(U; \phi) = \mathcal{N}(\mu_\phi, \Sigma_\phi)$ and mean-field Gaussians $q(z_n; \phi)$.
This is equivalent to maximizing the evidence lower bound (ELBO)~\cite{titsias2010bayesian}:
\begin{align}
    \mathcal{L} &= \sum_i \mathbb{E}_{q(z_{n_i}; \phi)\, q(f_i \mid z_{n_i}; W, \theta, \phi)} \left[ \log p(y_i | f_i) \right] - D_\text{KL}[q(U; \phi) || p(U; W, \theta)] - \sum_{n=1}^N D_\text{KL}[q(z_n; \phi) || p(z)]
\end{align}
wherein the outer expectation is approximated by Monte Carlo over the codes: only patient $n_i$'s code enters $f_i$, and given a sample $z_{n_i} \sim q(z_{n_i}; \phi)$---which fixes $\hat{x}_i$ through $\Circled{B}$---the inducing values marginalize analytically~\cite{hensman2015scalable},
\begin{align}
    q(f_i \mid z_{n_i}; W, \theta, \phi) &= \int p(f_i \mid U, \hat{x}_i; W, \theta) \, q(U; \phi) \, \mathrm{d}U \nonumber \\
    &= \mathcal{N}\left( \psi_i \cdot \mu_\phi, \; K_\theta(\hat{x}_i, \hat{x}_i) + \psi_i \cdot (\Sigma_\phi -  K_\theta(W, W)) \cdot \psi_i^\intercal \right),
\end{align}
with $\psi_i = K_\theta(\hat{x}_i, W) \cdot K_\theta(W, W)^{-1}$. The remaining one-dimensional expectation of $\log p(y_i \mid f_i)$ under this Gaussian has no closed form for a Bernoulli--sigmoid likelihood, and is itself estimated by Monte Carlo.
The first term of $\mathcal{L}$ can be estimated via mini-batching with just $O(M^3)$ per gradient step~\cite{hensman2015scalable,hensman2013gaussian,lalchand2022generalised}.

\textbf{Reading Patient Similarity from Learned Latent Representations.}
We can better understand patient similarity, even in high-dimensional latent spaces, by treating the model's covariance as a graph, provided that the kernel over $\hat{x}$ can be decomposed into a product of kernels applied to $x$ and $z$: $K_\theta(\widehat{X}, \widehat{X}') = K_\theta^x(X, X') \odot K_\theta^z(Z, Z')$.
Specifically, we compute the covariance matrix between \emph{patients} (not observations) by applying the latent-space kernel, $K_\theta^z$, to the means of the variational posteriors, $q(z_n; \phi)$.
We then treat this covariance as an adjacency matrix, in which patients are nodes and edge weights equal the covariance between the patients.

\textbf{Exploring and Interpreting Learned Similarity by Mapping to Patient Demographics.}
To deepen our understanding of patient similarity, we can now explore how it aligns with demographics or social determinants of health in two ways.
(1) Qualitatively: we visualize the similarity graph, pruning edges with covariances below a chosen threshold to reduce visual clutter.
We then color vertices by group (whether demographic or other), as well as edges connecting patients within the same group, keeping remaining edges black (see \cref{fig:graph}).
The relative number of colorful to black edges visually communicates how similar patients are to members of the same group.
(2) Quantitatively: we can compare similarity within vs.~between demographic groups using \emph{modularity}~\cite{newman2004finding}.
Modularity, $Q$, lies in $[-1/2, 1)$, where $Q = 0$ means within-group edges carry no more weight than the degree-preserving null model predicts. 
Originally proposed to identify community structures in networks, modularity has recently been used on covariance matrices, for example, to identify connectivity of brain regions (e.g.~\cite{yun2020brain,rakesh2023network}); here, we compute it on our similarity graph.
Given a covariance matrix, $K \in \mathbb{R}^{N \times N}$ (with diagonal entries set to $0$ to remove self-loops), and group membership $g_n$ for every patient $n$, $Q$ is computed via:
\begin{align}
    Q &= \frac{1}{\mathrm{vol}} \sum\limits_{n, n'} \left[ K_{n,n'} - \frac{\gamma}{\mathrm{vol}} \cdot \left( \sum\limits_i K_{i,n'} \right) \cdot \left( \sum\limits_j K_{n,j} \right) \right] \cdot \mathbb{I}(g_n = g_{n'}),
\end{align}
where $\mathrm{vol} = \sum_{n, n'} K_{n,n'}$ is the graph's ``volume'' (with $K$ symmetric this sums over ordered pairs, so it is twice the total edge weight) and $\gamma$ is the ``resolution parameter'' (commonly set to $1$), which scales the expected within-group weight under the degree-preserving null model---a larger $\gamma$ demands more within-group weight before $Q$ turns positive, biasing detection toward smaller communities.

\section{Experiments and Results: Measuring Forecasting Performance Under Patient Heterogeneity} \label{sec:results}

\begin{table*}[t]
\caption{\textbf{Comparison of Methods on Test Metrics.} We report log-likelihood by stratifying patients into bottom, middle, and top thirds based on number of data points. Idiographic models outperform their Single counterparts on every metric except specificity. LVMOGP matches the strongest baselines on most metrics; VB-LR remains best overall.}
\label{tab:results}

\tiny
\centering
\setlength{\tabcolsep}{3pt}

\begin{tabular}{l|l||cccc|c|c|c|c}
                             &  \textbf{Method}       & \multicolumn{4}{c}{\textbf{Avg. Log-Likelihood}}                                   & \textbf{ROC-AUC}         & \textbf{PPV}             & \textbf{Sensitivity}     & \textbf{Specificity}    \\
                             &   & \textit{Bottom}           & \textit{Middle}           & \textit{Top}              & \textit{All}              & \textit{All}             & \textit{All}             & \textit{All}             & \textit{All}            \\ \midrule 
\multirow{5}{*}{\rotatebox[origin=c]{90}{Single}}      
                             & \textbf{KNN}     & N/A              & N/A              & N/A              & N/A              & $0.70 \pm 0.01$  & $0.61 \pm 0.02$ & $0.22 \pm 0.02$ & $0.97 \pm 0.00$ \\
                             & \textbf{RBF-GP}      & $-0.62 \pm 0.01$ & $-0.49 \pm 0.01$ & $-0.36 \pm 0.00$  & $-0.43 \pm 0.00$  & $0.74 \pm 0.00$  & $0.66 \pm 0.04$ & $0.15 \pm 0.01$ & $0.98 \pm 0.00$ \\
                             & \textbf{LR}      & $-0.66 \pm 0.00$  & $-0.53 \pm 0.01$ & $-0.38 \pm 0.00$  & $-0.45 \pm 0.00$  & $0.68 \pm 0.01$ & $0.60 \pm 0.06$  & $0.05 \pm 0.01$ & $0.99 \pm 0.00$ \\
                             & \textbf{EB-LR}   & $-1.87 \pm 0.00$  & $-1.72 \pm 0.00$  & $-1.09 \pm 0.00$  & $-1.56 \pm 0.00$  & $0.68 \pm 0.01$ & $0.61 \pm 0.03$ & $0.05 \pm 0.01$ & $0.99 \pm 0.00$ \\
                             & \textbf{VB-LR}   & $-1.87 \pm 0.00$  & $-1.72 \pm 0.00$  & $-1.09 \pm 0.00$  & $-1.56 \pm 0.00$  & $0.68 \pm 0.01$ & $0.61 \pm 0.05$ & $0.05 \pm 0.01$ & $0.99 \pm 0.00$ \\ \midrule
\multirow{5}{*}{\rotatebox[origin=c]{90}{Idiographic}} 
                             & \textbf{KNN}     & N/A              & N/A              & N/A              & N/A              & $0.78 \pm 0.01$ & $0.73 \pm 0.02$ & $0.34 \pm 0.03$ & $0.97 \pm 0.00$ \\
                             & \textbf{RBF-GP}      & $-0.52 \pm 0.01$ & $-0.42 \pm 0.01$ & $-0.31 \pm 0.00$  & $-0.37 \pm 0.00$  & $0.84 \pm 0.00$  & $0.73 \pm 0.01$ & $0.37 \pm 0.01$ & $0.97 \pm 0.00$ \\
                             & \textbf{LR}      & $-0.49 \pm 0.01$ & $-0.40 \pm 0.01$  & $-0.29 \pm 0.01$ & $-0.34 \pm 0.01$ & $0.85 \pm 0.01$ & $0.73 \pm 0.01$ & $0.36 \pm 0.02$ & $0.97 \pm 0.00$ \\
                             & \textbf{EB-LR}   & $-0.54 \pm 0.01$ & $-0.51 \pm 0.00$  & $-0.38 \pm 0.01$ & $-0.43 \pm 0.00$  & $0.84 \pm 0.01$ & $0.71 \pm 0.02$ & $0.46 \pm 0.03$ & $0.96 \pm 0.00$ \\
                             & \textbf{VB-LR}   & $-0.47 \pm 0.01$ & $-0.38 \pm 0.01$ & $-0.28 \pm 0.01$ & $-0.33 \pm 0.01$ & $0.87 \pm 0.00$  & $0.73 \pm 0.01$ & $0.42 \pm 0.02$ & $0.96 \pm 0.00$ \\ \midrule
                             & \textbf{LVMOGP} & $-0.50 \pm 0.01$  & $-0.40 \pm 0.01$  & $-0.29 \pm 0.01$ & $-0.35 \pm 0.01$ & $0.85 \pm 0.01$ & $0.73 \pm 0.01$ & $0.37 \pm 0.02$ & $0.97 \pm 0.00$
\end{tabular}
\end{table*}

\textbf{Without significant kernel design, our approach already matches the strongest baselines on most metrics, showing promise.}
We compare our method to baselines in their ability to better forecast SREs one week in advance (details in \cref{apx:setup}).
\cref{tab:results} shows that, only having naively experimented with \emph{a single kernel}, our method already matches the best-performing baseline on nearly all metrics. 
We anticipate that a future investigation into the inductive biases of different kernels will allow our method to outperform all baselines, since LVMOGP generalizes both the GP and LR-based baselines.

\begin{figure*}[t]
    \centering

    \begin{subfigure}{0.49\textwidth}
    \centering
    \tiny
    Gender: Modularity $Q = 0.12$
    \end{subfigure}
    ~
    \begin{subfigure}{0.49\textwidth}
    \centering
    \tiny
    Sexual Orientation: Modularity $Q = 0.13$
    \end{subfigure}
    \begin{subfigure}{0.49\textwidth}
    \includegraphics[width=\linewidth]{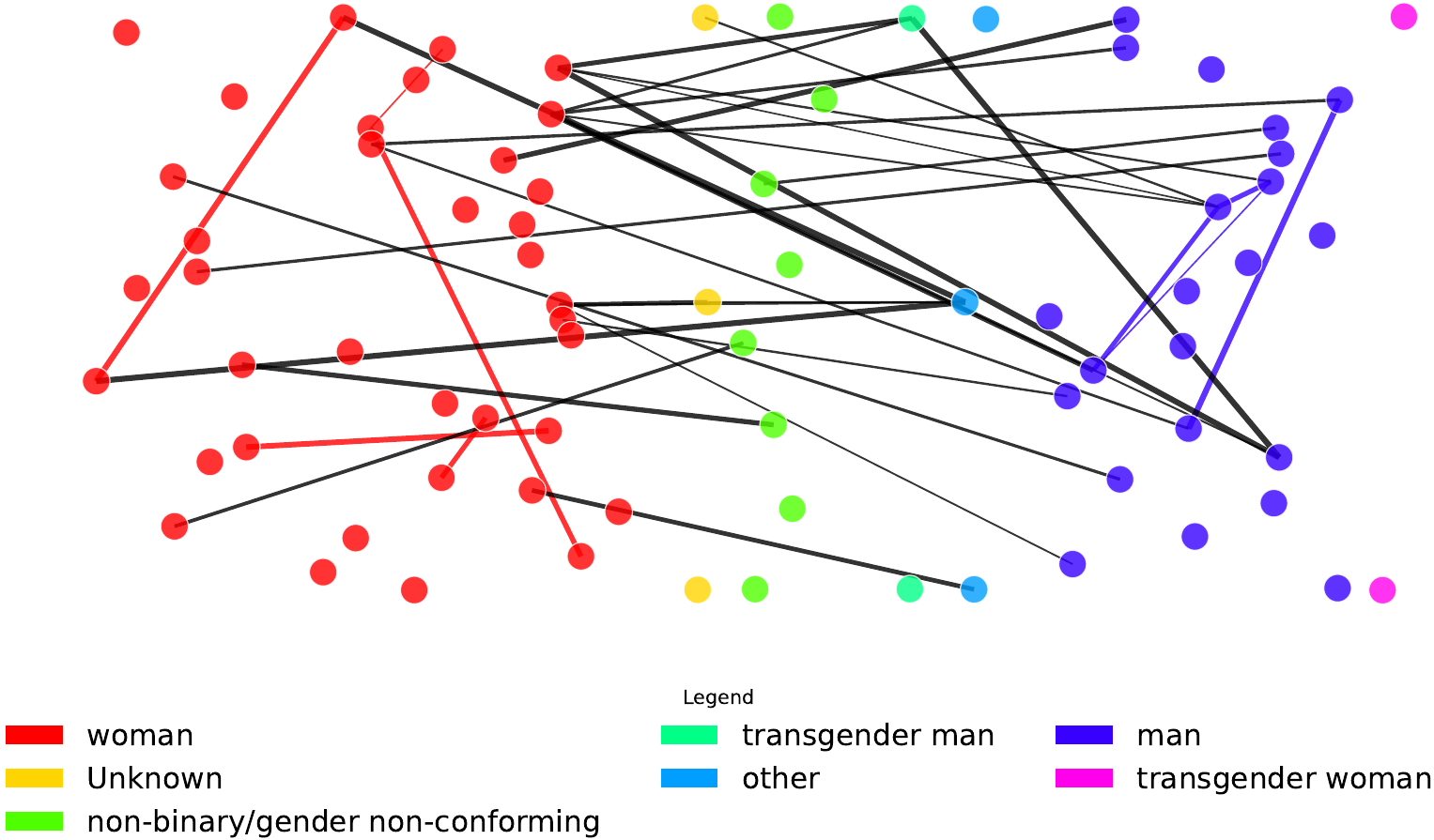}
    \end{subfigure}
    ~
    \begin{subfigure}{0.49\textwidth}
    \includegraphics[width=\linewidth]{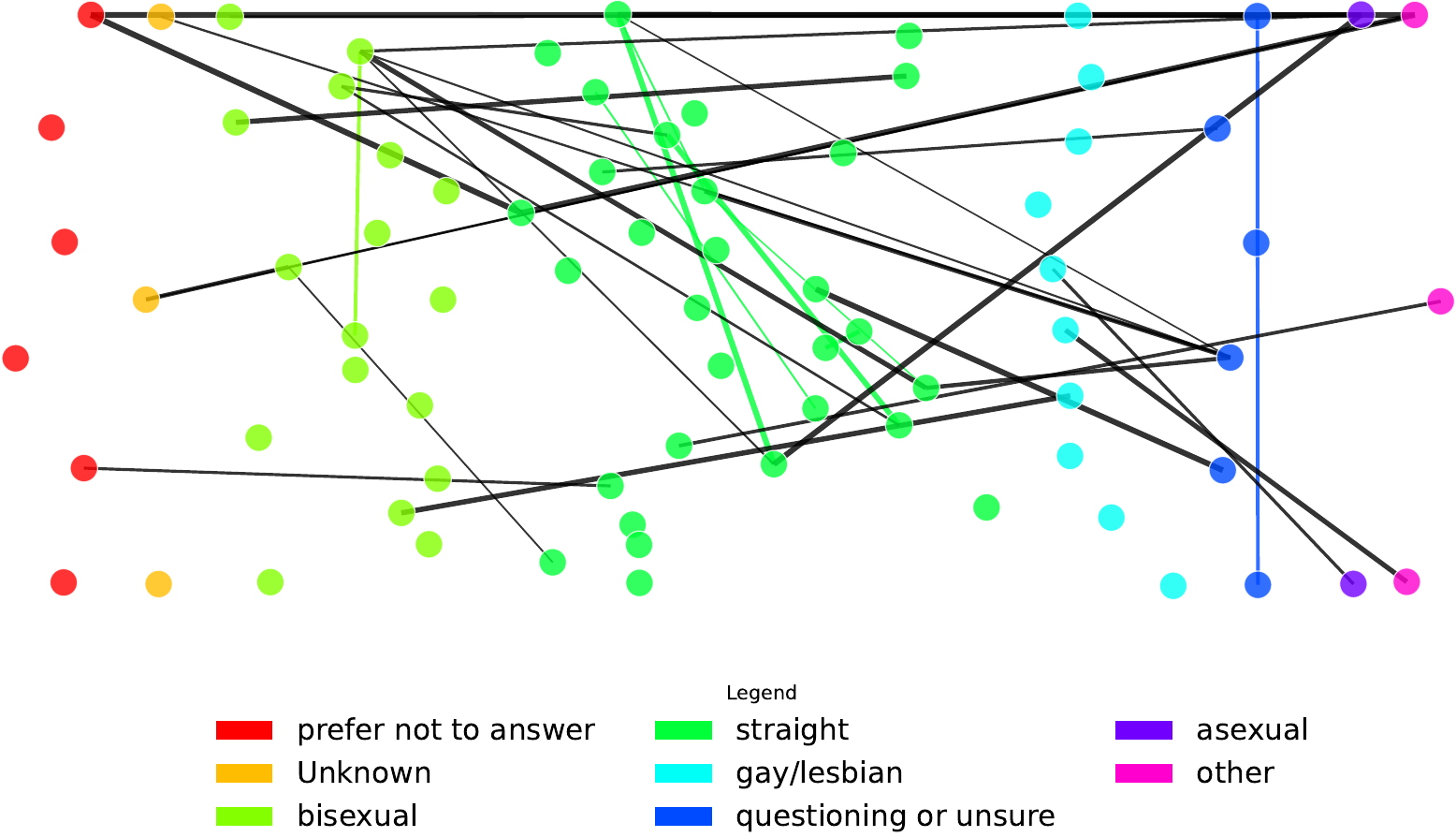}
    \end{subfigure}

    \vspace{2mm}

    \begin{subfigure}{0.49\textwidth}
    \centering
    \tiny
    Age: Modularity $Q = 0.08$
    \end{subfigure}
    ~
    \begin{subfigure}{0.49\textwidth}
    \centering
    \tiny
    Highest Completed Education: Modularity $Q = 0.15$
    \end{subfigure}
    
    \begin{subfigure}{0.49\textwidth}
    \includegraphics[width=\linewidth]{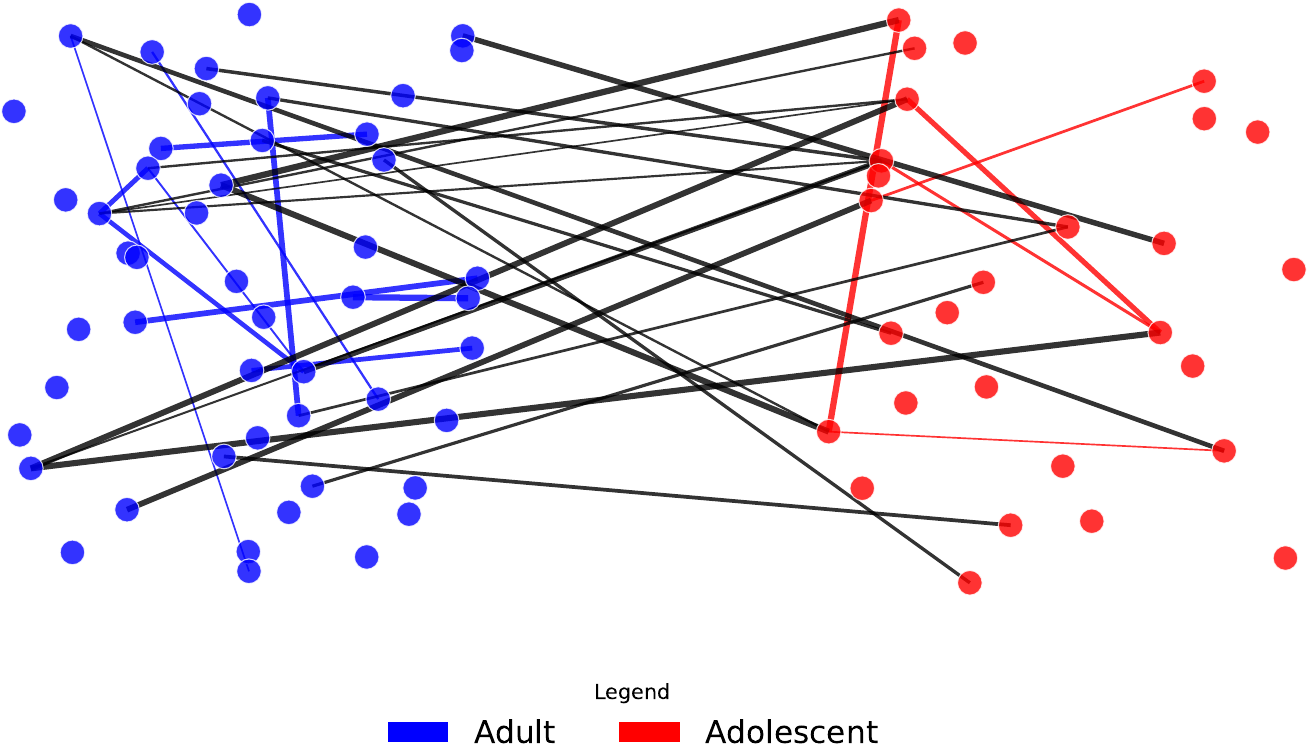}
    \end{subfigure}
    ~
    \begin{subfigure}{0.49\textwidth}
    \includegraphics[width=\linewidth]{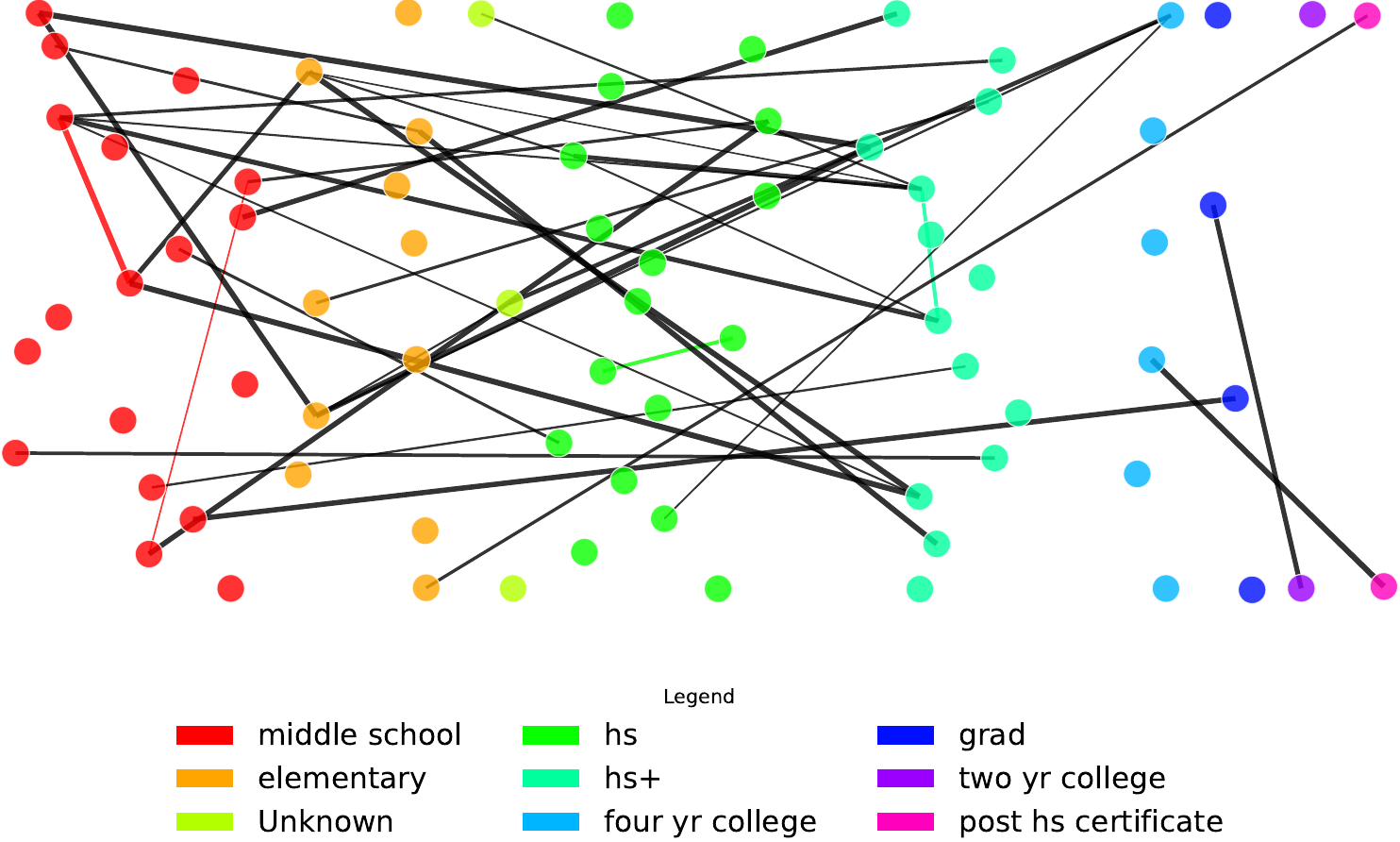}
    \end{subfigure}
    
    \caption{\textbf{Graphs of Patient Similarity.} Color of nodes represents group membership. Edges are black if connecting nodes of different groups; thickness indicates magnitude of covariance. Modularity $Q$ is reported per grouping. Group sizes are unequal and differ between panels, so each $Q$ should be read against the ceiling attainable for its own grouping rather than against $1$ (\cref{sec:method}).}
    \label{fig:graph}
\end{figure*}

\textbf{Insights from Patient Similarity Graphs.}
We visualize the similarity of patients in \cref{fig:graph} across different demographic groups, and find modularities between $0.08$ and $0.15$.
These are small in absolute terms, but as noted in \cref{sec:method}, a modularity is interpretable only against the ceiling attainable for its own grouping, which for unequal group sizes can be far below $1 - 1/G$; we do not compute those ceilings, nor a label-permutation null, here.
We therefore read this as preliminary evidence that these demographic factors do not account for patient similarity---the same direction as \cref{fig:random-groups-experiment}, in which random groupings outperform demographic ones---rather than as a measured effect size.
Interestingly, in nearly all graphs, the \emph{largest} similarities are between patients of different groups (mostly black edges remain after pruning).

\section{Limitations and Future Work} \label{sec:future}

Four key directions will strengthen our preliminary findings (\cref{sec:results}). The first asks whether our predictive claim survives a harder test---an unfiltered cohort, split temporally. The remaining three ask variations on the same question: can we truly capture patient similarity? That is, we ask whether the model determines it, whether our inference recovers it, and whether our summary of it can be read at face value. 

\paragraph{Towards a Deployable Model.} Our cohort keeps patients with at least three SREs and three non-SREs so that every split contains both classes, and our splits assign each patient's surveys without regard to when they occurred. Within this setting, our central observation stands: risk trajectories differ enough that idiographic models beat a single one. To deploy such a model, however, we require a different analysis, which we will pursue in future work. Conditioning the cohort on the outcome excludes patients who never event---the majority of any population the model would fit to. And because, in this setting, an SRE labels the whole prior week, one event spans a long run of positively labeled surveys, so a split that assigns them independently may place a single episode on both sides of the cut, flattering the idiographic models more than the single ones. In future work we therefore plan to fit the LVMOGP to an unfiltered cohort, split temporally into train/validation/test, with whole episodes confined to one split. This analysis asks a harder and more clinically meaningful question---can the heterogeneity we observe be \emph{exploited prospectively}?

\paragraph{Identifiability of Learned Patient Similarity.} Our input-space kernel reads the latent code (through $b_\theta, v_\theta, c_\theta$; \cref{apx:setup}), so patients can come to resemble one another in two places: in the latent kernel $K^z_\theta$, which is what we plot, and in the input kernel $K^x_\theta$, which is not. Therefore, it may be possible for similarity to ``leak out'' of the graph---the model may treat two patients as alike in a way that shapes both their forecasts, yet never draw an edge between them. In future work, we plan to further investigate the identifiability of the learned patient similarity.

\paragraph{Effects of Approximate Inference.} As in most work on GPs with latent variables, we take $q(U; \phi)$ independent of the codes, $Z$, and $q(Z; \phi)$ as mean-field across patients. In studying a similar class of models---Bayesian Neural Networks with Latent Variables---\citet{yacoby2022mitigating} show that explanatory power can be transferred between the global parameters and the latent variables while fitting the data equally well; as a consequence the posterior mode over both is asymptotically biased away from the ground truth, and standard mean-field inference yields parameters that generalize poorly and misestimate uncertainty. LVMOGP pairs global parameters with one latent variable per patient and is fit with exactly such a factorized approximation, so mean-field approximate inference may exhibit the same pathology here. We plan to investigate this in future work. 

\paragraph{Interpreting Similarity Graphs.} 
As noted in \cref{sec:results}, our reported modularities reflect raw values rather than calibrated effect sizes because we calculate neither grouping-specific ceilings nor label-permutation nulls. Properly interpreting Q requires contextualizing it against baseline modularity in size-matched random graphs~\cite{reichardt2006networks}.

\paragraph{Conclusion.} Forecasting suicide risk from EMA data requires capturing patient heterogeneity.
We show that single-trend models blur distinct risk trajectories, idiographic models overfit sparse data, and discrete demographic groupings miss underlying behavioral dynamics.
Modeling heterogeneity as a continuous latent variable via LVMOGP resolves these trade-offs, enabling data-sparse patients to share statistical strength while learning interpretable patient similarities.
We are optimistic that addressing key limitations---moving to prospective temporal evaluation, establishing kernel identifiability, and refining variational posterior approximations---will unlock the full potential of this framework and provide us with a rigorous instrument for personalized suicide risk forecasting.

\begin{ack}
Research reported in this publication was supported by the National Institute of Mental Health (NIMH) of the National Institutes of Health (NIH) under Award Numbers R01MH139486, U01MH116928, and P50MH132766. The content is solely the responsibility of the authors and does not necessarily represent the official views of the National Institutes of Health. The authors are also grateful for funding from the Tommy Fuss Collaborative STAR Fund, the Chet and Will Griswold Suicide Prevention Fund, and a gift from an anonymous donor. Finally, the authors are grateful to Wellesley College for supporting GH in the summer of 2025.
\end{ack}

\FloatBarrier

\medskip

{
\small
\bibliographystyle{unsrtnat}
\bibliography{references}
}


\appendix

\section{Overview of the EMA Data} \label{apx:data}

\textbf{Participants.}
A total of 623 unique participants presenting with suicidal thoughts and/or recent suicidal behavior were recruited from two hospitals in the Boston area---315 adults (ages 18+) from a psychiatric emergency service, and 308 adolescents (ages 12--19) from a psychiatric inpatient unit.
Participants were excluded if they did not own an iOS/Android smartphone or presented any of the following: a factor that impaired their ability to provide informed consent/assent, an inability to speak or write English fluently, a gross cognitive impairment due to florid psychosis, intellectual disability, dementia, acute intoxication, or extremely agitated or violent behavior. 

\textbf{Consent, Compensation, and IRB.}
After agreeing to participate, individuals signed consent/assent forms, answered an initial questionnaire, and installed the LifeData application on their mobile devices, which prompted them with brief self-report questionnaires. 
Participants received \$10 for completing the initial questionnaire and earned \$1 for each EMA survey they submitted. 
The study was approved by our institutions' IRB.

\textbf{Surveys.}
Smartphone surveys assessed participants' current experience of suicidal thinking---urge, intent, and ability to resist suicidal urges---as well as 16 affective states---negative, hopeless, trapped, isolated, burdensome, angry, self-hate, agitated, worried, numb, fatigued, humiliated, desire to escape, desire to avoid, energetic, and positive---on a 0--10 Likert scale.
These surveys were sent to participants 6 times per day for three months, with the first and last sent at fixed times decided in collaboration with each participant, and the remaining surveys sent at randomized times between the first and last surveys. 
In addition to these surveys, participants could always opt to fill in additional surveys, for example, to report a suicide attempt, non-suicidal self-injury, or another event they deemed important. 
The study was monitored by a risk-monitoring team in real time to intervene when participants indicated high suicidal intent (details available upon request).

\textbf{Recording SREs.}
An SRE was recorded in the data if it was reported by the patient in the survey, if it was reported by the risk-monitoring team, or if it was reported in the patient's electronic health record (consensus-coded by two trained BA-level reviewers with supervision by a doctoral-level clinician with expertise in assessing/treating STBs).

\textbf{Data Inclusion in Analysis.}
We kept all SREs for which there was at least one EMA survey in the week prior.
We kept data from all patients who had at least 3 SREs and 3 non-SREs to ensure we can include one of each in the train/validation/test split (see \cref{apx:setup}).
Due to the low base rate of SREs, this left us with $N = 75$ patients who contributed a total of $14794$ complete EMA surveys.

\section{Experimental Setup} \label{apx:setup}

\textbf{Data Splits.}
We divided the data into sets of size $50\%$, $25\%$, and $25\%$ for training, validation, and test, respectively.
We ensured that there was at least one SRE and one non-SRE in each set.
As such, we assume that for our method to be used in practice, patients must have at least one recorded SRE in their data.
We created these cuts of the data 5 times, conducting all experiments on each cut of the data, and reporting the mean $\pm$ standard deviation of all metrics.

\textbf{Random Restarts.}
For each cut of the data, we ran each method 5 times, each with a random seed.
We selected the best-performing random restart on the validation Log-Likelihood. 

\textbf{Hyperparameter Selection.} 
We performed a grid search over the following parameters, selecting them based on Log-Likelihood on the validation set:
\begin{itemize}
    \item \textbf{KNN:} Neighbors $k \in \{1, 2\}$, which performed best in our preliminary experiments, and distance $\in \{\text{Minkowski, Manhattan}\}$. We used the default parameters from \texttt{scikit-learn}~\cite{scikit-learn} for the remaining parameters.
    
    \item \textbf{LR:} Default parameters from \texttt{scikit-learn}~\cite{scikit-learn} but with a maximum of 5000 iterations until convergence.

    \item \textbf{VB-LR:} We trained for a maximum of 5000 iterations until convergence, with the shape and rate $\alpha, \beta$ on the Gamma prior on the precision of the coefficients both $\in \{ 1.0, 2.0 \}$, and with the remaining parameters set to the defaults from \citet{sklearnbayes}.

    \item \textbf{EB-LR:} We trained for a maximum of 5000 iterations, with the initial precision of the prior distribution $\alpha \in \{ 3.0, 2.0, 1.0, 1\mathrm{e}{-3}, 1\mathrm{e}{-6}, 1\mathrm{e}{-9}, 1\mathrm{e}{-12} \}$, and with the remaining parameters set to the defaults from \citet{sklearnbayes}.

    \item \textbf{GP:} We used the GP classifier from \texttt{scikit-learn}~\cite{scikit-learn} with its default hyperparameters; it places an isotropic RBF kernel over the inputs (a single shared lengthscale, rather than one per dimension) and approximates the posterior with a Laplace approximation. We additionally set $\texttt{max\_iter\_predict} = 5000$, as well as $\texttt{n\_restarts\_optimizer} = 1$, which selects across two kernel hyperparameter initializations---default and random. 
    
    \item \textbf{LVMOGP:} We used $M = 2000$ inducing points, $D_z = 3$, and a mini-batch size of $B = 150$.
    We fit the model with $15000$ gradient steps and a learning rate of $0.005$.
    Finally, we used a kernel that factorizes as $K_\theta(\widehat{X}, \widehat{X}') = K_\theta^x(X, X') \odot K_\theta^z(Z, Z')$, with $K_\theta^z$ as an ARD kernel and $K_\theta^x$ as a ``state-dependent'' linear kernel, defined by:
    \begin{align}
        k_\theta(\hat{x}, \hat{x}') &= b_\theta(z) \cdot b_\theta(z') + v_\theta(z) \cdot v_\theta(z') \cdot (x - c_\theta(z))^\intercal \cdot (x' - c_\theta(z')),
    \end{align}
    where $b_\theta(\cdot)$, $v_\theta(\cdot)$, and $c_\theta(\cdot)$ are each a neural network with parameters $\theta$. 
    This kernel allows patients to have differing priors depending on their location in the latent space.

\end{itemize}

\textbf{Software.}
We implemented the LVMOGP in NumPyro~\cite{bingham2019pyro} and JAX~\cite{jax2018github}.

\end{document}